\documentclass[letterpaper, 10 pt, conference]{packages/ieeeconf}

\IEEEoverridecommandlockouts                              

\usepackage{times}
\usepackage{epsfig}
\usepackage{graphicx}
\usepackage{amsmath}
\usepackage{amssymb}
\usepackage{subfigure} 
\usepackage[ruled]{algorithm}
\usepackage{algpseudocode}
\usepackage{ctable}
\usepackage[export]{adjustbox}
\usepackage[table]{xcolor}
\usepackage{import}
\usepackage{pgf}
\usepackage{cite}
\usepackage[font=small,skip=4pt]{caption}
\makeatletter
\let\NAT@parse\undefined
\makeatother
\usepackage[numbers,sort&compress]{natbib}

\usepackage{packages/sparo_acronyms}
\usepackage{packages/sparo_math}
\usepackage{packages/sparo_SIunits}
\usepackage{packages/sparo_misc}
\usepackage{multirow}
\usepackage{hyperref}

\usepackage{soul,color}
\usepackage{verbatim} 
\usepackage{makecell} 

\usepackage{xcolor}

\definecolor{gl}{HTML}{008000}

\usepackage[symbol]{footmisc}

\title{\LARGE \bf
    RadLoc: Radar-based 3-DoF Global Localization via Fast, Robust, and Lightweight Spatial Descriptor Across Diverse Environmental Scenarios \vspace{-0.2cm} 
}

\author{Hogyun Kim$^{1}$, Jiwon Choi$^{1}$, Jungwoo Lee$^{1}$, and Younggun Cho$^{1\dagger}$
\thanks{$^{1}$Hogyun Kim, $^{1}$Jiwon Choi, $^{1}$Jungwoo Lee, and $^{1\dagger}$Younggun Cho are with the Electrical Engineering, Inha University,
Incheon, South Korea
{\tt\small [hg.kim, jiwon2, pihsdneirf]@inha.edu, yg.cho@inha.ac.kr}  \hfill \break
This work was supported by National Research Foundation of Korea (NRF) grant (No. RS-2026-25551458 and RS-2025-02217000) and Institute of Information \& communications Technology Planning \& Evaluation (IITP) grant (RS-2022-II220448) funded by the Korea government~(MSIT).%
}
\vspace{-0.15cm}
}

\begin{document}

\maketitle
\thispagestyle{empty}
\pagestyle{empty}

\begin{abstract}
While global localization using spinning radar has gained attention for its robustness to adverse weather and challenging environments, many studies have focused on individual components such as place recognition or pose estimation. 
In this paper, we take a holistic view of radar sensor-based global localization and present \textit{RadLoc}, a fast, robust, and lightweight end-to-end pipeline from place recognition to 3-DoF pose estimation. 
RadLoc accelerates pre-processing using 1D CA-CFAR filtering and leverages the near-range dominance in spinning radar images to design a compact descriptor and an efficient hierarchical coarse-to-fine retrieval strategy. 
Moreover, coupled with phase correlation-based 3-DoF pose estimation, it forms a versatile global localization module applicable to SLAM and multi-session SLAM systems. 
Extensive experiments on 15 sequences across 5 datasets demonstrate that RadLoc achieves robust performance while maintaining the smallest descriptor size and fastest retrieval time among state-of-the-art approaches.
The supplementary materials are available at \url{https://sparolab.github.io/research/radloc/}.
\end{abstract} 
\section{Introduction}
Global localization, which recognizes revisited places (\textit{i.e.,} place recognition) and estimates the relative pose between them, is a fundamental capability for autonomous mobile robots, underpinning applications such as simultaneous localization and mapping~(SLAM)~\cite{adolfsson2023tbv} and multi-session/multi-robot mapping~\cite{kim2026commerge, kim2025skid}.
Unlike vision and light detection and ranging~(LiDAR) sensor-based approaches that suffer from short-wavelength attenuation under adverse weather, radio detection and ranging~(radar) sensors penetrate rain, fog, snow, and dust due to their longer wavelength.
For this reason, radar sensor-based global localization has been actively studied and has demonstrated its effectiveness across diverse environments and weather conditions~\cite{hong2022radarslam}.

Despite remarkable progress~\cite{kim2020mulran, jang2023raplace, gadd2024open, kim2024referee, choi2024referee} over the past decade, prior works have often focused on enhancing the performance of individual components~(\textit{e.g.,} place recognition or pose estimation), disregarding computational efficiency and the integration into an overall global localization framework.
In particular, while recent learning-based approaches~\cite{suaftescu2020kidnapped, kim2025sherloc} have shown strong recognition performance, they often require extensive training data and substantial computational resources, limiting their deployability in new environments. 

\begin{figure}[t]
    \centering
    \def\width{0.49\textwidth}%
    \includegraphics[clip, trim={10 0 0 0}, width=0.49\textwidth]{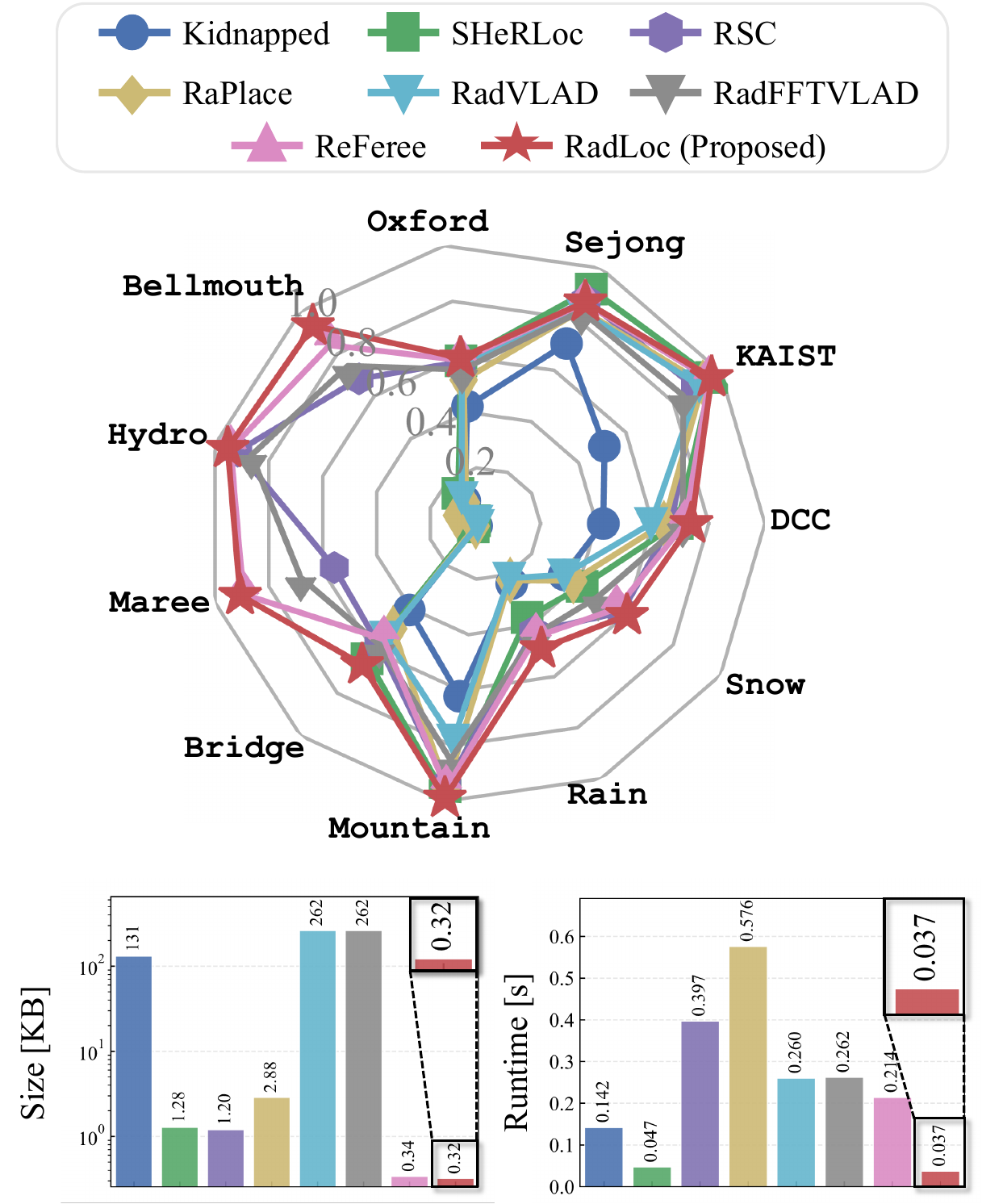}
    \caption{Comparison of Recall@1~(top), descriptor size~(bottom-left), and retrieval runtime~(bottom-right) with state-of-the-art approaches~\cite{suaftescu2020kidnapped, kim2025sherloc, kim2020mulran, jang2023raplace, gadd2024open, kim2024referee} on 11 sequences across 5 datasets~\cite{kim2020mulran, barnes2020oxford, gadd2024oord, kim2025hercules, burnett2023boreas}. 
    Despite variations in environments, weather conditions, and radar types, our \textit{RadLoc} consistently demonstrates robust recall performance while achieving the smallest descriptor size and the fastest place recognition time~(\textit{i.e.,} description generation time and retrieval time) among all compared methods.} 
    \label{fig:main} 
    \vspace{-0.5cm}
\end{figure}

While prior works primarily improve individual components, we instead adopt a holistic perspective on radar sensor-based global localization by jointly considering descriptor design, retrieval efficiency, and pose estimation within a unified framework.
As shown in \figref{fig:main}, in this paper, we propose \textit{RadLoc}, a fast, robust, and lightweight global localization pipeline.

Specifically, we first accelerate pre-processing by replacing costly feature extraction~\cite{cen2018precise} with a fast 1D cell-averaging constant false alarm rate~(CA-CFAR)~\cite{rohling2011ordered} filtering along the range direction.
Next, we leverage the near-range dominance in spinning radar images to design a range-aware descriptor and an efficient hierarchical coarse-to-fine retrieval scheme. 
Finally, combined with phase correlation-based 3-DoF pose estimation, our RadLoc forms an end-to-end global localization pipeline.

Extensive experiments on 15 sequences across 5 datasets demonstrate that RadLoc achieves robust recognition across diverse environments and radar types, while maintaining the smallest descriptor size and the fastest retrieval runtime among state-of-the-art methods. 
Moreover, the proposed pipeline naturally extends to SLAM, multi-session SLAM, and cross-weather map management scenarios.
\section{Related Works}
In this section, we review prior works on global localization with LiDAR and radar sensors. 
Rather than enumerating individual methods, we analyze their common design patterns to clarify the gap that RadLoc addresses.

\subsection{LiDAR sensor-based Global Localization}
LiDAR sensor-based global localization has been extensively studied, particularly through bird's-eye view~(BEV) representations of 3D point clouds. 
Scan Context~\cite{kim2018scan} pioneered a compact descriptor by encoding a 3D scan into a 2D polar BEV image, enabling efficient place recognition with 1-DoF yaw estimation via column-wise circular shifting. 
Scan Context++~\cite{kim2021scan} extended this idea to improve robustness against lateral variations while maintaining rotational alignment.
Beyond handcrafted descriptors, RING~\cite{lu2022one} and RING++~\cite{xu2023ring++} employed the Radon transform to construct rotation- and translation-invariant global descriptors from BEV images, estimating translation via cross-correlation and yaw via circular shifting. 
Learning-based approaches such as BEVPlace~\cite{luo2023bevplace} further leveraged neural feature extraction on BEV representations to achieve robust recognition with 2-DoF position estimation, and BEVPlace++~\cite{luo2025bevplace++} extended this to full 3-DoF pose estimation using rotation-equivariant local features and keypoint matching.

Collectively, these methods share a common paradigm: projecting high-dimensional spatial measurements onto structured 2D representations, performing efficient place retrieval, and subsequently estimating relative pose via correlation or feature matching. 
This structured design pattern has proved effective for LiDAR sensor-based global localization.

\subsection{Radar sensor-based Global Localization}
Radar sensor-based global localization has primarily focused on place recognition. 
Early works directly adapted LiDAR-inspired descriptors to scanning radar images. 
Radar Scan Context~\cite{kim2020mulran} transferred handcrafted polar descriptors, Kidnapped Radar~\cite{suaftescu2020kidnapped} employed metric learning to improve discriminability, and 
RadVLAD~\cite{gadd2024open} adopted VLAD-based aggregation on polar radar representations for efficient place retrieval.
However, raw radar images are inherently affected by receiver saturation, multipath reflections, and speckle noise, which limit descriptor robustness.

Subsequent studies introduced signal-level filtering and frequency-domain processing to mitigate these effects. 
RaPlace~\cite{jang2023raplace} utilized the Radon transform with adaptive thresholding for rotation- and translation-invariant matching, and RadFFTVLAD~\cite{gadd2024open} applied frequency-domain filtering to enhance invariance properties. 
ReFeree~\cite{kim2024referee} combined radar feature extraction with binary representation-based descriptors to achieve robust place recognition with 1-DoF yaw estimation. 
However, its reliance on costly feature extraction during pre-processing limits computational efficiency, and its pose estimation remains restricted to yaw alignment.

More recently, learning-based approaches such as SHeRLoc~\cite{kim2025sherloc} addressed cross-modal radar matching by learning discriminative embeddings that implicitly suppress noise and dynamic objects. 
However, such learning-based methods require extensive training data and substantial computational resources, limiting their deployability in new environments.

Consequently, there remains a need for a lightweight and unified framework that integrates efficient retrieval with full 3-DoF pose estimation, as summarized in~\tabref{tab:related_works}.

\begin{table}[t]
\captionsetup{width=0.49\textwidth, justification=justified} 
\caption{Summary of related works on LiDAR and radar sensor-based global localization, highlighting the scope of pose estimation. 
Unlike LiDAR sensor-based methods that have progressed toward full 3-DoF pose estimation, existing radar sensor-based approaches remain limited to place recognition or 1-DoF yaw estimation.}
\setlength{\tabcolsep}{4pt} 
\centering\resizebox{0.49\textwidth}{!}{\scriptsize
\renewcommand{\arraystretch}{1.2}
\begin{tabular}{l|l|l|l|l}
\toprule \midrule
Sensor                  & Approach                             & Venue           & Year  & Description \\ \midrule
\multirow{8}{*}{\hspace{4pt}\rotatebox[origin=r]{90}{LiDAR}} & Scan Context \cite{kim2018scan}      & IROS            & 2018  & Place Recognition + 1-DoF Yaw Estimation  \\ 
                        & Scan Context++ \cite{kim2021scan}    & TRO             & 2022  & Place Recognition + 1-DoF Yaw Estimation \\ \cmidrule{2-5}
                        & RING \cite{lu2022one}                & IROS            & 2022  & Place Recognition + 3-DoF Pose Estimation \\ 
                        & RING++ \cite{xu2023ring++}           & TRO             & 2023  & Place Recognition + 3-DoF Pose Estimation \\ \cmidrule{2-5}
                        & BEVPlace \cite{luo2023bevplace}      & ICCV            & 2023  & Place Recognition + 2-DoF Position Estimation \\ 
                        & BEVPlace++ \cite{luo2025bevplace++}  & TRO             & 2025  & Place Recognition + 3-DoF Pose Estimation \\ \cmidrule{2-5}
                        & SOLiD \cite{kim2024narrowing}        & RA-L            & 2024  & Place Recognition + 1-DoF Yaw Estimation \\ \midrule
\multirow{2}{*}{\hspace{4pt}\rotatebox[origin=r]{90}{Radar}}   & ReFeree \cite{kim2024referee}        & RA-L            & 2024  & Place Recognition + 1-DoF Yaw Estimation  \\ \cmidrule{2-5}
                        & RadLoc (Proposed)                    & ---             & 2026  & Place Recognition + 3-DoF Pose Estimation  \\ \midrule
\bottomrule
\end{tabular}}
\label{tab:related_works}
\vspace{-0.5cm}
\end{table}

\begin{figure}[b]
    \vspace{-0.5cm}
    \centering
    \def\width{0.49\textwidth}%
    \includegraphics[clip, trim={0 0 0 0}, width=0.49\textwidth]{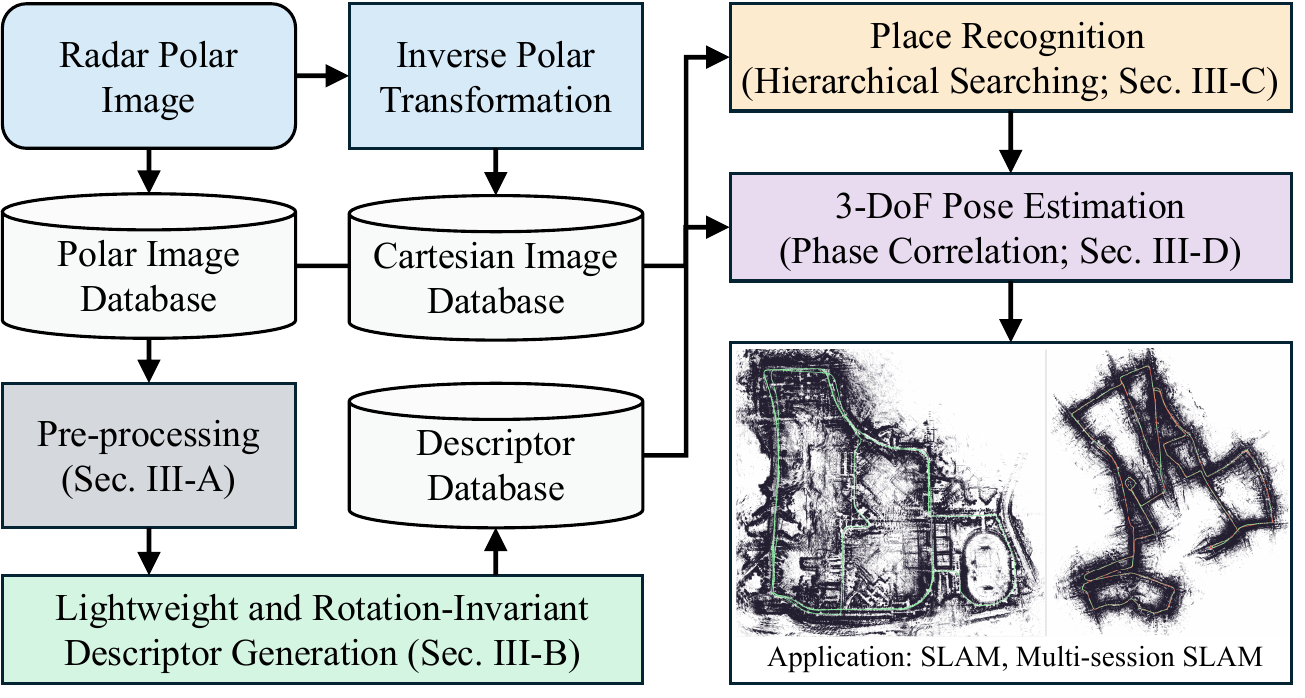}
    \caption{Overview of the proposed \textit{RadLoc} pipeline. 
             Through fast pre-processing~(Sec.~\ref{sec:preprocessing}), lightweight and rotation-invariant descriptor generation~(Sec.~\ref{sec:descriptor_generation}), efficient hierarchical place recognition~(Sec.~\ref{sec:place_recognition}), and robust 3-DoF pose estimation~(Sec.~\ref{sec:pose_estimation}), we present RadLoc applicable to various tasks such as SLAM and multi-session SLAM.}
    \label{fig:pipeline}
\end{figure}

\section{Fast, Robust, and Lightweight Radar-based Global Localization}
An overview of the \textit{RadLoc} pipeline is summarized in \figref{fig:pipeline}.
Given a polar radar image from a spinning radar sensor, we convert it into a Cartesian representation for pose estimation while retaining the original polar image for descriptor construction.
We then apply lightweight pre-processing to suppress radar-specific noise~(Sec.~\ref{sec:preprocessing}), and the processed image is used to construct a compact descriptor, which is stored in a database for efficient retrieval~(Sec.~\ref{sec:descriptor_generation}). 
At query time, RadLoc performs hierarchical coarse-to-fine place retrieval against the database~(Sec.~\ref{sec:place_recognition}), and retrieves the corresponding polar and Cartesian images of the matched place for 3-DoF pose estimation via phase correlation~(Sec.~\ref{sec:pose_estimation}).
In the following subsections, we detail each component.

\subsection{Fast Radar Image Pre-processing}
\label{sec:preprocessing}
First, let $\mathbf{I}_{\text{polar}} \in \mathbb{R}^{N_a \times N_r}$ denote a polar radar image, where $N_a$ is the number of azimuth bins and $N_r$ is the number of range bins.
Each element $p_{ij} = \mathbf{I}_{\text{polar}}{(i,j)}$ represents the received power intensity at azimuth index $i\in\{1,\cdots,N_a\}$ and range index $j\in\{1,\cdots,N_r\}$.
Here, since the radar performs a full $360^\circ$ azimuth sweep with a known maximum range $r_{\text{max}}$, the metric-scale azimuth and range of each bin $(i,j)$ can be determined as $\theta_i = \frac{2\pi(i-1)}{N_a}$, $r_j = j\cdot \Delta r$, where $\Delta r = \frac{r_{\text{max}}}{N_r}$ denotes the range resolution.

Raw radar images are inherently corrupted by various sources of noise, including speckle noise or receiver saturation, which degrade the quality of global localization~\cite{yang2023noise}.
To suppress these, existing methods typically rely on feature extraction techniques such as \cite{cen2018precise}.
In addition, to improve robustness against radar noise, \citet{kim2024referee} leverage binary representation rather than occupancy, where $f_{ij} \in \{0, 1\}$ is binarized from the raw intensity $p_{ij}$ via feature extraction.

\begin{figure}[t]
    \centering
    \def\width{0.49\textwidth}%
    \includegraphics[clip, trim={20 0 0 15}, width=0.49\textwidth]{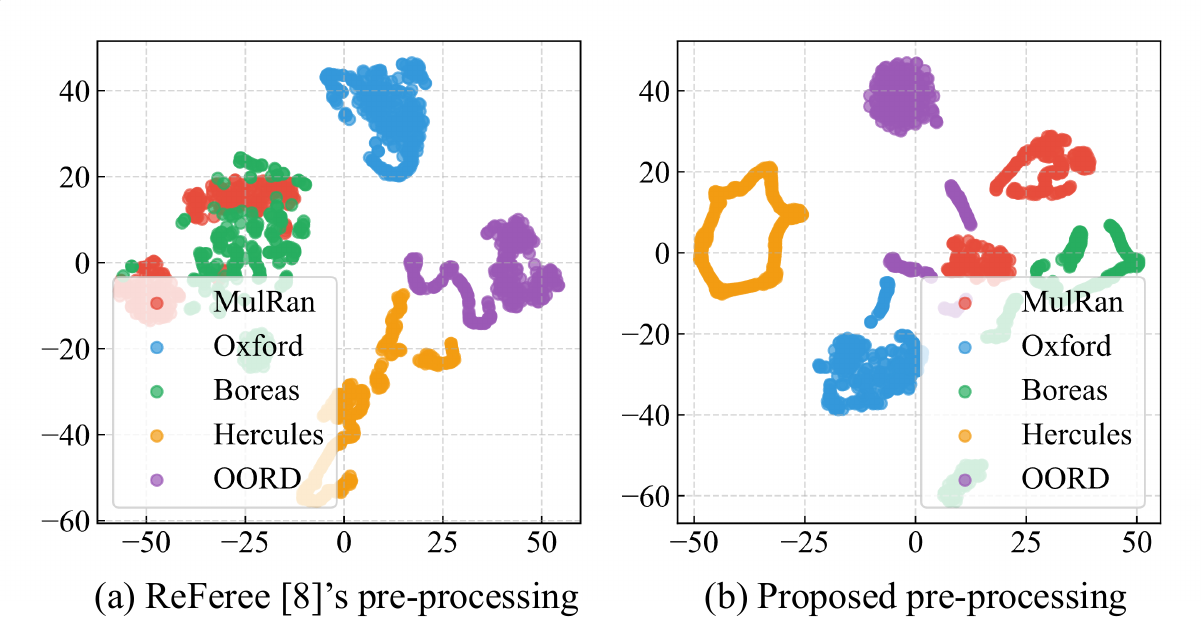}
    \caption{t-SNE visualization of descriptors generated by the proposed pipeline with (a)~conventional radar feature- and binary representation-based pre-processing~\cite{kim2024referee} and (b)~our proposed pre-processing~(Sec.~III-A). 
    Our pre-processing improves cluster separability, as indicated by higher Silhouette scores ~(0.359 than 0.330) and Calinski–Harabasz scores ~(8339.89 than 963.56) across 5 datasets~\cite{kim2020mulran, barnes2020oxford, burnett2023boreas, kim2025hercules, gadd2024oord}.}
    \label{fig:tsne}
    \vspace{-0.5cm}
\end{figure}

However, such feature extraction methods incur computational overhead, and the binary representation is sensitive to the choice of extraction parameters, which may vary across radar types and environments~(\figref{fig:tsne}(a)).
To address these issues, we replace costly feature extraction with a lightweight 1D CA-CFAR filtering~\cite{rohling2011ordered} along the range direction, and retain its response as a continuous intensity representation.
Together, utilizing 1D CA-CFAR and continuous intensity representation makes our pre-processing both fast and more discriminative~(\figref{fig:tsne}(b)).

\subsection{Compact and Rotation-Invariant Descriptor Generation}
\label{sec:descriptor_generation}
Next, let $\tilde{\mathbf{I}}_{\text{polar}} \in \mathbb{R}^{N_a \times N_r}$ denote the pre-processed polar radar image, where each element $\tilde{p}_{ij}$ represents the preprocessed intensity.
From this representation, we construct a compact descriptor that is both robust and efficient for place retrieval.
To this end, we apply average pooling with patch size $(N_a, p_r)$ over the pre-processed polar image, aggregating the full azimuth sweep into a single value per range patch as:
\begin{equation}
    d_n = \frac{1}{N_a \cdot p_r} 
    \sum_{i=1}^{N_a} 
    \sum_{j=(n-1)p_r+1}^{n \cdot p_r} \tilde{p}_{ij},
    \quad n \in \{1, \cdots, E\},
\end{equation}
where $E = \frac{N_r}{p_r}$ is the descriptor dimension.
The resulting elements are then concatenated into a compact 1D descriptor as follows:
\begin{equation}
    \mathbf{d} = [d_1, d_2, \cdots, d_{E}]^\top 
    \in \mathbb{R}^{E}.
\label{eq:desc}
\end{equation}

This design offers two key advantages. 
First, aggregation over the full azimuth dimension eliminates dependency on yaw alignment, yielding inherent rotational invariance. 
Second, the range pooling factor $p_r$ directly controls the descriptor dimensionality, allowing a compact representation even for long-range spinning radars and enabling efficient large-scale retrieval.
For instance, in the MulRan dataset~\cite{kim2020mulran}, where $N_r = 3360$, setting $p_r = 84$ reduces the descriptor dimension to only $E=40$, substantially lowering retrieval cost.

\subsection{Simple yet Effective Hierarchical Searching}
\label{sec:place_recognition}
The spatial information density of polar radar measurements decreases with increasing range.
As the range increases, received power decays according to the radar equation, and each polar bin spans a larger Cartesian area, leading to reduced signal strength and structural resolution.
As a result, discriminative structural cues concentrate in the near-to-mid range, while far-range responses become progressively less reliable.

Importantly, the proposed descriptor preserves this radial ordering by construction.
Since each descriptor element corresponds to a contiguous range segment after azimuth aggregation, the descriptor forms a radial structural profile of the scene, explicitly ordered from near- to far-range.
This explicit radial structure naturally enables a hierarchical coarse-to-fine retrieval strategy: the coarse stage focuses on the informative near-to-mid range, while the fine stage verifies matches using the full descriptor.

\begin{figure}[t]
    \centering
    \def\width{0.49\textwidth}%
    \includegraphics[clip, trim={0 0 0 0}, width=0.49\textwidth]{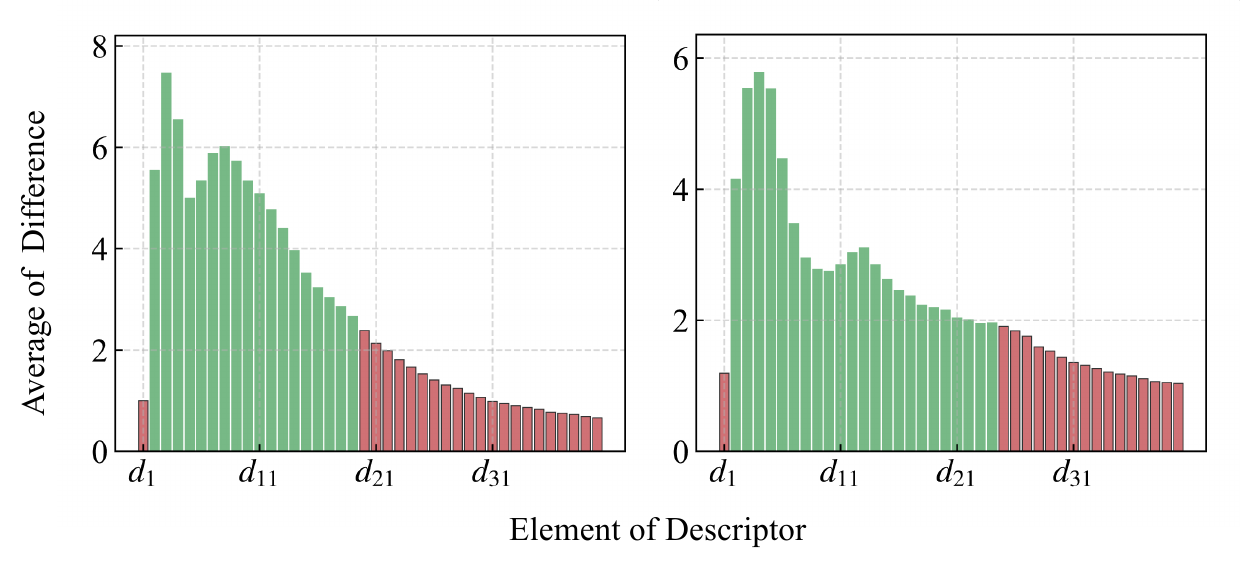}
    \caption{Average of difference per descriptor element for negative loop pairs~(distance $>$ 20\,m) on the (left)~snowy day, \textit{i.e.,} \texttt{Mountain 03}, in Hercules~\cite{kim2025hercules} and (right)~sunny day, \textit{i.e.,} \texttt{21-01-19}, in Boreas~\cite{burnett2023boreas} datasets. 
    The 40 elements are ordered from near-range~($d_1$) to far-range~($d_{40}$), with green and red indicating elements below and above the third quartile~($Q_3$, 75\,\%) of the cumulative descriptor distance~(approximately around $d_{20}$), respectively. 
    }
    \label{fig:diff}
    \vspace{-0.5cm}
\end{figure}

Building upon this, we partition each descriptor into a near-range segment and the full descriptor as follows:
\begin{equation}
\mathbf{d}_{\text{near}} = [d_1, \cdots, d_K], \qquad
\mathbf{d}_{\text{full}} = [d_1, \cdots, d_{E}],
\end{equation}
where $K$ is a user-defined partition index.
Based on the analysis of descriptor element contributions (Fig.~\ref{fig:diff}), we set $K=20$ in this study.
Furthermore, while the first element exhibits a relatively small magnitude, it captures the immediate surrounding structure that is sensitive to abrupt local changes (\textit{e.g.,} occlusions or nearby dynamic objects) and is therefore retained.

\textbf{Coarse stage.}
Within the near-range segment, we observe a gradual decrease in contribution with range, which is largely attributable to range-dependent attenuation rather than the absence of structural information.
In typical outdoor scenes, mid-range regions (\textit{e.g.,} tens of meters) may still contain stable and distinctive structures such as building facades, but their responses can be under-represented due to power decay.
To compensate for this attenuation and reduce range-dependent bias within the near-range segment, we apply a monotonically increasing weighting as follows:
\begin{equation}
\tilde{\mathbf{d}}_{\text{near}} = \mathbf{w} \odot \mathbf{d}_{\text{near}},
\end{equation}
where $\mathbf{w} = [w_1, \ldots, w_K]^\top$ is a user-defined range-proportional model and $\odot$ denotes the Hadamard product.
While the observed difference reflects a genuine range-dependent characteristic of radar measurements, our goal is not to eliminate this effect but to moderate its dominance.
In practice, we adopt a simple monotonic weighting $w_k = \sqrt{k}$, where $k\in\{1,\cdots,K\}$, to mildly compensate for range-dependent attenuation.
This formulation provided stable retrieval performance without over-amplifying far-range responses.

Using these weighted near-range descriptors~$\{\tilde{\mathbf{d}}_{\text{near}}^{(1)}, \cdots, \tilde{\mathbf{d}}_{\text{near}}^{(N_{\text{db}})}\}$, where $N_{\text{db}}$ denotes the number of descriptors stored in the database, we first construct a Kd-tree index $\mathcal{T}_{\text{near}}$.
For a query descriptor $\mathbf{d}^{(q)}$, its weighted near-range component $\tilde{\mathbf{d}}_{\text{near}}^{(q)}$ is computed and used to retrieve the top-$N_c$ nearest neighbors from $\mathcal{T}_{\text{near}}$ by minimizing the $L_2$ distance as follows:
\begin{equation}
s_{\text{coarse}}(m) =
\left\|
\tilde{\mathbf{d}}_{\text{near}}^{(q)} -
\tilde{\mathbf{d}}_{\text{near}}^{(m)}
\right\|_2,
\quad m \in \mathcal{I}_{\text{db}},
\end{equation}
where $\mathcal{I}_{\text{db}} = \{1,\cdots,N_{\text{db}}\}$.
The resulting candidate set is denoted as $\mathcal{C} \subset \mathcal{I}_{\text{db}}$ with $|\mathcal{C}| = N_c$.
This Kd-tree-based coarse retrieval substantially reduces the search space with logarithmic query complexity. 

As illustrated in Fig.~\ref{fig:diff}, the average absolute difference at the $d_3$ is approximately $7.5$, whereas $d_{20}$ is about $2.1$. 
Under an $L_2$ distance, this leads to a squared contribution ratio of $(7.5/2.1)^2 \approx 12.8\times$, indicating strong near-range dominance. 
In contrast, a fully normalized $L_2$ metric would collapse this ratio toward $1\times$, removing scale disparity altogether.
With the proposed weighting ($w_3=\sqrt{3}$, $w_{20}=\sqrt{20}$), the squared contribution ratio becomes $(\sqrt3 \cdot 7.5)^2/(\sqrt{20} \cdot 2.1)^2 \approx 1.9\times$, positioning our method between these two extremes and achieving a balanced attenuation of range-dependent bias while preserving structural contrast.

\textbf{Fine stage.}
The top-$N_c$ candidates are then re-ranked using the full descriptor without range-dependent weighting as follows:
\begin{equation}
s_{\text{fine}}(m) =
\left\|
\mathbf{d}_{\text{full}}^{(q)} -
\mathbf{d}_{\text{full}}^{(m)}
\right\|_2,
\quad m \in \mathcal{C}.
\end{equation}
Here, the top-ranked candidate after re-ranking is selected as the final retrieval result. 
We intentionally avoid applying range-dependent weighting in this stage, since emphasizing far-range bins may amplify low signal-to-noise ratio~(SNR) and degrade matching stability.
Using the original unweighted descriptor yields more reliable matching across the entire range extent.

\subsection{3-DoF Pose Estimation via Phase Correlation}
\label{sec:pose_estimation}
Given the top-ranked candidate retrieved from the hierarchical search, we estimate the relative 3-DoF pose~$(\Delta x, \Delta y, \Delta \theta)$ between the query and the matched radar image using phase correlation.
Phase correlation is based on the Fourier shift theorem: if two images differ by a translational shift, their cross-power spectrum exhibits a distinct peak at the corresponding offset.
By leveraging the Fourier shift theorem in appropriate coordinate domains, rotation and translation can be estimated via peak detection in the cross-power spectrum.
Inspired by~\cite{park2020pharao}, we decompose 3-DoF pose estimation into two sequential steps: rotation estimation in the log-polar domain, followed by translation estimation in the Cartesian domain.

\textbf{Rotation estimation.} 
Let $\mathbf{I}_{\text{cart}}^{(q)}$ and $\mathbf{I}_{\text{cart}}^{(m)}$ denote the Cartesian radar images of the query and the matched candidate, respectively. 
Since each bin in the Cartesian image corresponds to a known metric-scale position derived from the radar's azimuth and range resolution (Sec.~\ref{sec:preprocessing}), the resulting pose estimates are inherently metric-scaled. 
The Cartesian images are converted into log-polar coordinates, in which a planar rotation corresponds to a horizontal translational shift. 
By applying phase correlation on the log-polar representations, we directly obtain the metric-scale rotational offset $\Delta \theta$.

\textbf{Translation estimation.}
After estimating the rotational offset $\Delta \theta$, we first compensate for this rotation in the polar domain by circularly shifting the polar image along the azimuth axis.
The rotation-compensated polar image is then converted back into the Cartesian coordinate system to obtain $\mathbf{I}_{\text{cart}}^{(m, \text{rot})}$.
We subsequently apply 2D phase correlation between $\mathbf{I}_{\text{cart}}^{(q)}$ and the rotation-aligned $\mathbf{I}_{\text{cart}}^{(m, \text{rot})}$ to recover the metric-scale translational offset $(\Delta x, \Delta y)$.

This two-stage decomposition avoids coupling between rotation and translation, leading to stable and efficient 3-DoF pose estimation.
\definecolor{myemerald}{rgb}{0.753, 0.898, 0.804}
\definecolor{mylightgreen}{rgb}{0.894, 0.933, 0.745}
\definecolor{myyellow}{rgb}{0.996, 0.972, 0.780}
\definecolor{mygray}{gray}{0.90}
\newcommand{\firstc}{\cellcolor{myemerald!100}}
\newcommand{\secondc}{\cellcolor{mylightgreen!100}}
\newcommand{\thirdc}{\cellcolor{myyellow!100}}
\newcommand{\grayc}{\cellcolor{mygray}}

\newcommand{\sunny}{\raisebox{-0.5ex}{\includegraphics[height=2.2ex]{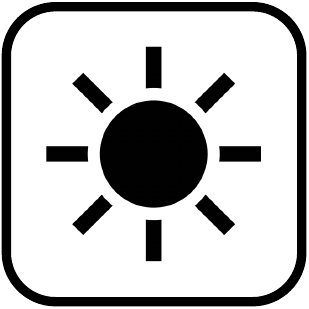}}}
\newcommand{\cloud}{\raisebox{-0.5ex}{\includegraphics[height=2.2ex]{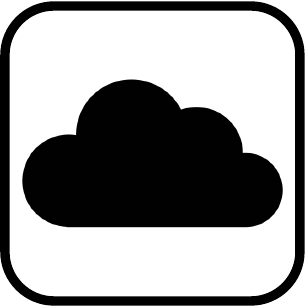}}}
\newcommand{\snow}{\raisebox{-0.5ex}{\includegraphics[height=2.2ex]{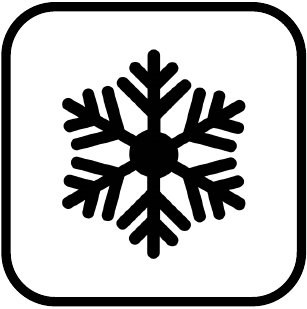}}}
\newcommand{\rain}{\raisebox{-0.5ex}{\includegraphics[height=2.2ex]{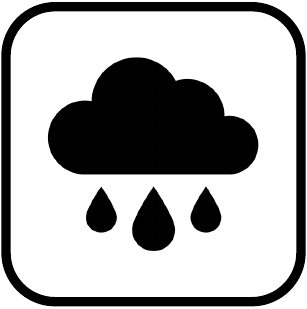}}}

\begin{table*}[t]
\caption{Single-session place recognition performance compared with state-of-the-art approaches~\cite{suaftescu2020kidnapped, kim2025sherloc, kim2020mulran, jang2023raplace, gadd2024open, kim2024referee}. R@1 is evaluated under 10\,m and 20\,m loop closure thresholds across 15 sequences from 5 datasets~\cite{barnes2020oxford, gadd2024oord, kim2025hercules, kim2020mulran, burnett2023boreas}.}
\label{tab:single_session}
\renewcommand{\arraystretch}{1.2}
\centering
\resizebox{\textwidth}{!}{\scriptsize
\begin{tabular}{l|cccccccccccccccccc}
\toprule \midrule
Datasets & \multicolumn{6}{c}{Oxford Radar Robotcar \cite{barnes2020oxford} (\,$\sunny$\,)}            
                              & \multicolumn{6}{c}{OORD \cite{gadd2024oord} (\,$\sunny$\,)} 
                              & \multicolumn{6}{c}{Hercules \cite{kim2025hercules} (\,$\sunny$ $\cloud$ $\snow$\,)}                      \\ 
\cmidrule(lr){2-7} \cmidrule(lr){8-13} \cmidrule(lr){14-19}
Sequence & \multicolumn{2}{c}{\texttt{19-01-10}}  
         & \multicolumn{2}{c}{\texttt{19-01-14}}  
         & \multicolumn{2}{c}{\texttt{19-01-18}}        
         & \multicolumn{2}{c}{\texttt{Bellmouth}} 
         & \multicolumn{2}{c}{\texttt{Hydro}}     
         & \multicolumn{2}{c}{\texttt{Maree}}            
         & \multicolumn{2}{c}{\texttt{Bridge 02}} 
         & \multicolumn{2}{c}{\texttt{Mountain 02}}     
         & \multicolumn{2}{c}{\texttt{Parking 01}}  \\ 

\cmidrule(lr){2-3} \cmidrule(lr){4-5} \cmidrule(lr){6-7} 
\cmidrule(lr){8-9} \cmidrule(lr){10-11} \cmidrule(lr){12-13}
\cmidrule(lr){14-15} \cmidrule(lr){16-17} \cmidrule(lr){18-19}

Evaluation
& \makecell{R@1\\{\tiny(10\,m)}} & \makecell{R@1\\{\tiny(20\,m)}}  
& \makecell{R@1\\{\tiny(10\,m)}} & \makecell{R@1\\{\tiny(20\,m)}}  
& \makecell{R@1\\{\tiny(10\,m)}} & \makecell{R@1\\{\tiny(20\,m)}}  
& \makecell{R@1\\{\tiny(10\,m)}} & \makecell{R@1\\{\tiny(20\,m)}}  
& \makecell{R@1\\{\tiny(10\,m)}} & \makecell{R@1\\{\tiny(20\,m)}}  
& \makecell{R@1\\{\tiny(10\,m)}} & \makecell{R@1\\{\tiny(20\,m)}}  
& \makecell{R@1\\{\tiny(10\,m)}} & \makecell{R@1\\{\tiny(20\,m)}}  
& \makecell{R@1\\{\tiny(10\,m)}} & \makecell{R@1\\{\tiny(20\,m)}}  
& \makecell{R@1\\{\tiny(10\,m)}} & \makecell{R@1\\{\tiny(20\,m)}}  \\ 
\midrule
Kidnapped \cite{suaftescu2020kidnapped} & 0.458                  & 0.422                  & 0.373                  & 0.341                  & 0.433                  & 0.422 
                             & 0.105                  & 0.107                  & 0.022                  & 0.047                  & 0.025                  & 0.029 
                             & 0.413                  & 0.408                  & 0.562                  & 0.622                  & 0.597                  & 0.671 
                             \\     
SHeRLoc \cite{kim2025sherloc}& \secondc 0.610         & \thirdc 0.586          & 0.575                  & 0.522                  & 0.662                  & \secondc 0.639 
                             & 0.146                  & 0.143                  & 0.014                  & 0.036                  & 0.037                  & 0.041 
                             & \secondc 0.580         & \secondc 0.638         & \secondc 0.979         & \secondc 0.979         & 0.801                  & 0.771 
                             \\
RSC \cite{kim2020mulran}& 0.589                  & 0.582                  & 0.566                  & 0.522                  & 0.635                  & 0.619 
                             & 0.638                  & 0.682                  & \thirdc 0.684          & \thirdc 0.916          & 0.448                  & 0.557 
                             & \thirdc 0.553          & \thirdc 0.574          & 0.932                  & 0.939                  & \firstc \textbf{0.936} & \secondc 0.832 
                             \\     
RaPlace \cite{jang2023raplace} & 0.554                  & 0.517                  & 0.531                  & 0.460                  & 0.571                  & 0.538 
                             & 0.099                  & 0.099                  & 0.063                  & 0.102                  & 0.031                  & 0.032 
                             & 0.511                  & 0.496                  & 0.954                  & 0.951                  & 0.746                  & 0.693 
                             \\     
RadVLAD \cite{gadd2024open}  & 0.591                  & 0.567                  & \thirdc 0.601          & \thirdc 0.533          & \thirdc 0.660          & \secondc 0.639 
                             & 0.134                  & 0.123                  & 0.013                  & 0.030                  & 0.020                  & 0.024 
                             & 0.500                  & 0.535                  & 0.770                  & 0.779                  & 0.627                  & 0.618 
                             \\     
RadFFTVLAD \cite{gadd2024open} & 0.578                  & 0.554                  & 0.518                  & 0.468                  & 0.608                  & 0.579 
                               & \thirdc 0.684          & \thirdc 0.743          & 0.640                  & 0.865                  & \thirdc 0.557          & \thirdc 0.682 
                               & 0.539                  & 0.571                  & 0.830                  & 0.850                  & 0.831                  & 0.757 
                             \\      
ReFeree \cite{kim2024referee}                     & \thirdc 0.606          & \secondc 0.588         & \firstc \textbf{0.604} & \secondc 0.542         & \secondc 0.663         & 0.633 
                             & \secondc 0.834         & \secondc 0.841         & \secondc 0.690         & \secondc 0.942         & \secondc 0.698         & \secondc 0.895 
                             & 0.538                  & 0.546                  & \thirdc 0.955          & \thirdc 0.954          & \thirdc 0.898          & \thirdc 0.825 
                             \\ 
RadLoc (Proposed)            & \firstc \textbf{0.618} & \firstc \textbf{0.597} & \firstc \textbf{0.604} & \firstc \textbf{0.549} & \firstc \textbf{0.676} & \firstc \textbf{0.642} 
                             & \firstc \textbf{0.934} & \firstc \textbf{0.933} & \firstc \textbf{0.706} & \firstc \textbf{0.953} & \firstc \textbf{0.757} & \firstc \textbf{0.907} 
                             & \firstc \textbf{0.640} & \firstc \textbf{0.665} & \firstc \textbf{0.991} & \firstc \textbf{0.988} & \secondc 0.907         & \firstc \textbf{0.846} 
\\ \midrule
\bottomrule
\end{tabular}
}

\vspace{2pt}

\begin{minipage}[t]{0.7\textwidth}
\vspace{0pt} 
\centering
\resizebox{\linewidth}{!}{\scriptsize
\begin{tabular}{l|cccccccccccc}
\toprule \midrule
Datasets & \multicolumn{6}{c}{MulRan \cite{kim2020mulran} (\,$\sunny$\,)}            
         & \multicolumn{6}{c}{Boreas \cite{burnett2023boreas} (\,$\sunny$ $\rain$ $\snow$\,)} \\ 
\cmidrule(lr){2-7} \cmidrule(lr){8-13}

Sequence & \multicolumn{2}{c}{\texttt{DCC 01}}  
         & \multicolumn{2}{c}{\texttt{KAIST 03}}  
         & \multicolumn{2}{c}{\texttt{Sejong 01}}        
         & \multicolumn{2}{c}{\texttt{21-01-19}} 
         & \multicolumn{2}{c}{\texttt{21-01-26}}     
         & \multicolumn{2}{c}{\texttt{21-04-29}} \\ 
\cmidrule(lr){2-3} \cmidrule(lr){4-5} \cmidrule(lr){6-7} 
\cmidrule(lr){8-9} \cmidrule(lr){10-11} \cmidrule(lr){12-13}
Evaluation
& \makecell{R@1\\{\tiny(10\,m)}} & \makecell{R@1\\{\tiny(20\,m)}}  
& \makecell{R@1\\{\tiny(10\,m)}} & \makecell{R@1\\{\tiny(20\,m)}}  
& \makecell{R@1\\{\tiny(10\,m)}} & \makecell{R@1\\{\tiny(20\,m)}}  
& \makecell{R@1\\{\tiny(10\,m)}} & \makecell{R@1\\{\tiny(20\,m)}}  
& \makecell{R@1\\{\tiny(10\,m)}} & \makecell{R@1\\{\tiny(20\,m)}}  
& \makecell{R@1\\{\tiny(10\,m)}} & \makecell{R@1\\{\tiny(20\,m)}}  \\ 
\midrule
Kidnapped \cite{suaftescu2020kidnapped}                    & 0.390                  & 0.424                  & 0.462                  & 0.509                  & 0.765                  & 0.705  
                             & 0.111                  & 0.140                  & 0.312                  & 0.338                  & 0.202                  & 0.235 \\     
SHeRLoc \cite{kim2025sherloc}                     & \firstc \textbf{0.711} & 0.688                  & \firstc \textbf{0.974} & \firstc \textbf{0.963} & \firstc \textbf{0.990} & \firstc \textbf{0.918}
                             & 0.166                  & 0.235                  & 0.384                  & 0.414                  & 0.288                  & 0.368 \\
RSC \cite{kim2020mulran}                    & 0.642                  & 0.667                  & 0.901                  & 0.895                  & 0.943                  & \secondc 0.874
                             & \secondc 0.416         & \secondc 0.447         & \secondc 0.555         & \secondc 0.583         & \thirdc 0.394          & 0.424\\     
RaPlace \cite{jang2023raplace}                     & 0.655                  & 0.640                  & 0.944                  & 0.926                  & 0.910                  & 0.837
                             & 0.163                  & 0.166                  & 0.378                  & 0.377                  & 0.219                  & 0.222 \\     
RadVLAD \cite{gadd2024open}                     & 0.607                  & 0.594                  & 0.922                  & 0.910                  & 0.906                  & 0.833
                             & 0.110                  & 0.137                  & 0.323                  & 0.338                  & 0.203                  & 0.214 \\     
RadFFTVLAD \cite{gadd2024open}                  & \secondc 0.692         & \thirdc 0.707          & 0.851                  & 0.843                  & 0.908                  & 0.832
                             & 0.273                  & 0.309                  & 0.469                  & 0.482                  & 0.381                  & \thirdc 0.436 \\      
ReFeree \cite{kim2024referee}                     & 0.689                  & \secondc 0.715         & \thirdc 0.962          & \thirdc 0.953          & \thirdc 0.943          & 0.864
                             & \thirdc 0.392          & \thirdc 0.443          & \thirdc 0.545          & \thirdc 0.561          & \secondc 0.406         & \secondc 0.443   \\ 
RadLoc (Proposed)                         & \thirdc 0.690          & \firstc \textbf{0.738} & \secondc 0.972         & \firstc \textbf{0.963} & \secondc 0.948         & \thirdc 0.868
                             & \firstc \textbf{0.454} & \firstc \textbf{0.499} & \firstc \textbf{0.597} & \firstc \textbf{0.605} & \firstc \textbf{0.464} & \firstc \textbf{0.490} \\
\midrule
\bottomrule
\end{tabular}
}
\end{minipage}
\hfill
\begin{minipage}[t]{0.28\textwidth}
\vspace{0pt} 
\centering
\includegraphics[clip, trim={10, 0, 0, 10}, width=\linewidth]{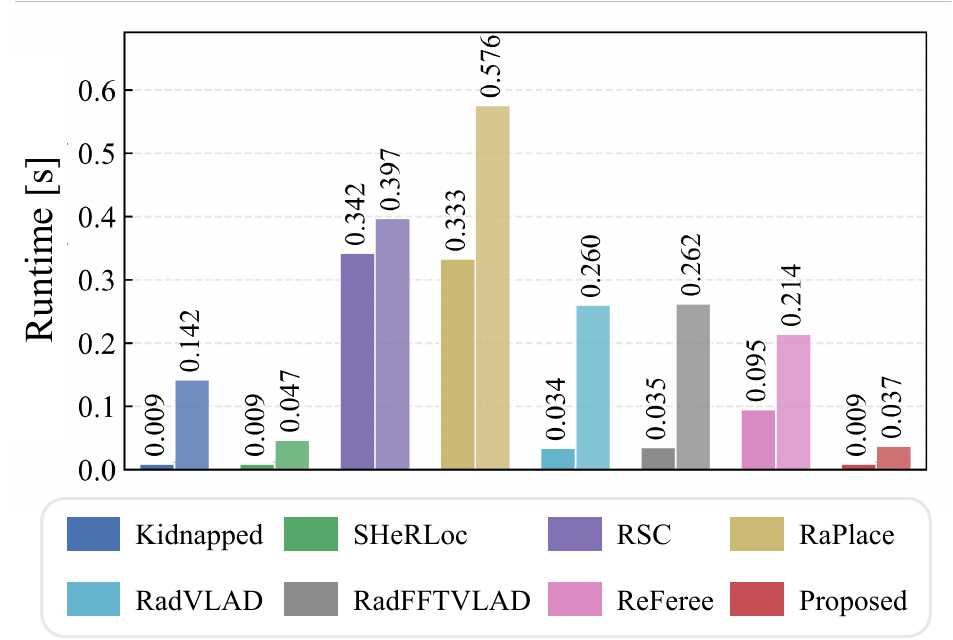}
\vspace{-0.35cm}
\captionof{figure}{Place recognition runtime comparison on the smallest~(\texttt{Parking 01}) to the largest~(\texttt{Maree}) dataset.}
\label{fig:pr_runtime}
\end{minipage}
\vspace{-0.7cm}
\end{table*}

\section{Experiments}
This section evaluates the effectiveness and efficiency of the proposed method for radar sensor-based global localization.
We demonstrate that our RadLoc (i) achieves robust place recognition across diverse datasets and weather conditions, (ii) enables accurate and efficient 3-DoF pose estimation for SLAM, and (iii) provides better scalability and runtime performance compared to state-of-the-art methods.

\subsection{Experimental Setup}
First, we evaluate our RadLoc on five public radar datasets: Oxford Radar Robotcar~\cite{barnes2020oxford}~$(\sunny)$, OORD~\cite{gadd2024oord}~$(\sunny)$, MulRan~\cite{kim2020mulran}~$(\sunny)$, Boreas~\cite{burnett2023boreas}~$(\sunny, \rain, \snow)$, and Hercules~\cite{kim2025hercules}~$(\sunny, \cloud, \snow)$.
These datasets span structured urban environments and unstructured off-road terrains, multiple weather conditions, and three different Navtech radar sensors, providing a comprehensive benchmark for radar sensor-based global localization.
Note that weather conditions are indicated using icons: \sunny~(sunny), \cloud~(cloudy), \rain~(rainy), and \snow~(snowy).

Next, we compare the proposed method against state-of-the-art learning-based and handcrafted approaches.
SHeRLoc~\cite{kim2025sherloc} and Kidnapped Radar~\cite{suaftescu2020kidnapped} employ deep learning, whereas Radar Scan Context (RSC)~\cite{kim2020mulran}, RaPlace~\cite{jang2023raplace}, RadVLAD~\cite{gadd2024open}, RadFFTVLAD~\cite{gadd2024open}, and ReFeree~\cite{kim2024referee} are handcrafted descriptor-based methods.
For 3-DoF pose estimation, we additionally compare against a feature extraction with generalized iterative closest point~(GICP) registration pipeline~\cite{kim2024referee}, serving as a strong feature-based baseline.

In the following, we organize the evaluation into three categories.
For place recognition~(Sec.~\ref{sec:pr}), we report descriptor generation and retrieval runtime, Recall@1 under 10\,m and 20\,m loop thresholds, area under the precision-recall curve (AUC), and maximum F1 score (F1).
For 3-DoF global localization~(Sec.~\ref{sec:pose_est}), we report relative translation error (RTE), relative rotation error (RRE), success rate, and pose estimation runtime.
A pose is considered successful if the estimated translation and rotation errors are within 5\,m and 10$^\circ$, respectively.
For ablation studies~(Sec.~\ref{sec:ablation}), we evaluate the proposed range-dependent weighting and hierarchical retrieval using the Kolmogorov–Smirnov~(KS) statistic, precision-recall curves, and runtime comparison.

For clarity, we highlight the top three results in each table as {\setlength{\fboxsep}{1pt}\colorbox{myemerald}{\textbf{1st}}}, {\setlength{\fboxsep}{1pt}\colorbox{mylightgreen}{2nd}}, and {\setlength{\fboxsep}{1pt}\colorbox{myyellow}{3rd}}.

\begin{table*}[t]
  \centering
  \begin{minipage}[c]{0.7\textwidth}
    \captionof{table}{Multi-session place recognition performance on the OORD~\cite{gadd2024oord}. R@1, AUC, and F1 are reported for three sequences.} 
    \label{tab:multi_session}
    \centering
    \renewcommand{\arraystretch}{1.2}
    \centering\resizebox{\textwidth}{!}{\tiny
    \begin{tabular}{l|ccccccccccccccc}
    \toprule \midrule
    Sequence   & \multicolumn{3}{c}{\texttt{Bellmouth}}      & \multicolumn{3}{c}{\texttt{Hydro}}         & \multicolumn{3}{c}{\texttt{Maree}} 
                                 \\ \cmidrule(lr){2-4} \cmidrule(lr){5-7} \cmidrule(lr){8-10}
    Evaluation & R@1 & AUC  & F1   & R@1 & AUC  & F1 & R@1 & AUC  & F1      \\ \midrule
    Kidnapped \cite{suaftescu2020kidnapped}                  & 0.008                  & 0.009                  & 0.020                  & 0.008                  & 0.006 
                                 & 0.013                  & 0.003                  & 0.002                  & 0.006
                                 \\     
    SHeRLoc \cite{kim2025sherloc}                    & \thirdc 0.928          & \secondc 0.867         & \secondc 0.861         & \firstc \textbf{0.989} & \secondc 0.949 
                                 & \secondc 0.915         & \secondc 0.981         & \secondc 0.857         & \firstc \textbf{0.868} 
                                 \\ 
    RSC \cite{kim2020mulran}                   & 0.772                  & \thirdc 0.860          & 0.761                  & 0.954                  & \thirdc 0.921 
                                 & 0.881                  & 0.634                  & 0.709                  & 0.626 
                                 \\     
    RaPlace \cite{jang2023raplace}                    & 0.467                  & 0.198                  & 0.439                  & 0.799                  & 0.386 
                                 & 0.652                  & 0.351                  & 0.114                  & 0.321 
                                 \\     
    RadVLAD \cite{gadd2024open}                    & 0.010                  & 0.011                  & 0.023                  & 0.006                  & 0.007 
                                 & 0.015                  & 0.005                  & 0.005                  & 0.010 
                                 \\     
    RadFFTVLAD \cite{gadd2024open}                & 0.915                  & 0.721                  & 0.798                  & 0.932                  & 0.684 
                                 & 0.771                  & 0.854                  & 0.463                  & 0.664 
                                 \\     
    ReFeree \cite{kim2024referee}                    & \secondc 0.950         & 0.803                  & \thirdc 0.852          & \secondc 0.984         & 0.895 
                                 & \thirdc 0.888          & \thirdc 0.968          & \thirdc 0.821          & \thirdc 0.813
                                 \\ 
    RadLoc (Proposed)            & \firstc \textbf{0.989} & \firstc \textbf{0.964} & \firstc \textbf{0.907} & \thirdc 0.980          & \firstc \textbf{0.956} 
                                 & \firstc \textbf{0.921} & \firstc \textbf{0.990} & \firstc \textbf{0.858} & \secondc 0.832 
    \\ \midrule 
    \bottomrule
    \end{tabular}
    }
  \end{minipage}%
  \hfill
  \begin{minipage}[c]{0.28\textwidth}
    \centering
    \includegraphics[clip, trim={0, 0, 0, 60}, width=\textwidth]{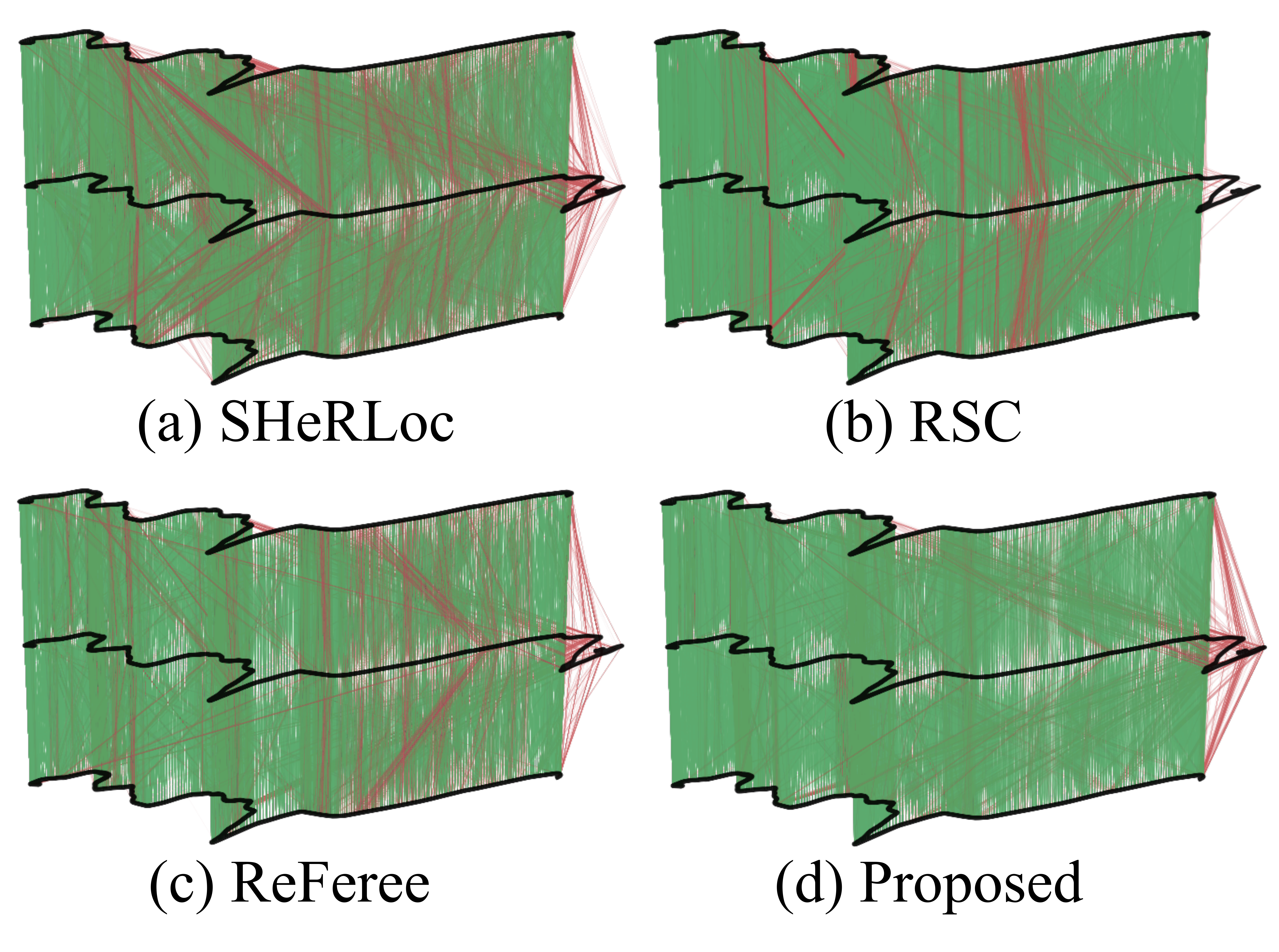}
    \vspace{-0.35cm}
    \captionof{figure}{Qualitative results of multi-session place recognition on three \texttt{Hydro} sequences for the top-4 methods.
    Green and red lines indicate true and false loop closures. 
    }
    \label{fig:pr_multi_session}
  \end{minipage}
    \vspace{-0.6cm}
\end{table*}

\subsection{Place Recognition Performance}
\label{sec:pr}
We first analyze the computational efficiency of our RadLoc, followed by its recognition performance in single-session and multi-session settings.

\textbf{Runtime analysis.}
As shown in \figref{fig:pr_runtime}, our RadLoc achieves the fastest total runtime among all compared methods on both the smallest~(\texttt{Parking 01}) and the largest~(\texttt{Maree}) datasets.
The runtime advantage becomes increasingly pronounced as the database size grows, confirming the scalability of the proposed hierarchical retrieval strategy.
Note that this efficiency makes our RadLoc well-suited for large-scale and real-time global localization applications.

\textbf{Single-session evaluation.}
Next, as reported in \tabref{tab:single_session}, our RadLoc consistently achieves top-tier R@1 performance across 15 sequences spanning 5 datasets, multiple radar types, and diverse weather conditions.
In unstructured off-road trajectories such as OORD, the proposed method outperforms competing methods, demonstrating robustness under reverse traversal and sparse geometric landmarks.
In structured urban environments (MulRan and Hercules datasets), where dense loop candidates increase ambiguity, our RadLoc remains among the top-performing methods while effectively suppressing false positives.
Notably, under adverse weather conditions in Boreas (\textit{e.g.,} \texttt{21-01-26}, \texttt{21-04-29}), our RadLoc maintains stable recall despite weather-induced intensity attenuation.

\textbf{Multi-session evaluation.}
Furthermore, as shown in \tabref{tab:multi_session} and \figref{fig:pr_multi_session}, several methods that perform competitively in single-session evaluation exhibit noticeable degradation when evaluated across sessions.
SHeRLoc shows improved multi-session performance associated with the same direction traversals across sessions.
Nevertheless, our RadLoc achieves the highest overall and most consistent results.

\subsection{3-DoF Global Localization}
\label{sec:pose_est}
Next, we evaluate our RadLoc for full 3-DoF global localization, including pose estimation accuracy, SLAM integration, and multi-session alignment under long-term variations.

\textbf{Pose estimation accuracy.}
As reported in \tabref{tab:regi_eval}, our RadLoc achieves competitive RTE and RRE across all evaluated datasets.
In particular, our RadLoc maintains a high success rate under the 5\,m and 10$^\circ$ threshold criteria, demonstrating reliable pose initialization even in large-scale and perceptually ambiguous environments.
Compared to the feature~\cite{cen2018precise}-based GICP alignment, our RadLoc provides more stable initialization while substantially reducing computational overhead.

\begin{table}[t]
\captionsetup{width=0.49\textwidth, justification=justified} 
\caption{3-DoF pose estimation performance on the Hercules dataset~\cite{kim2025hercules}. Each environment includes sequences captured under varying weather~(\textit{e.g.,} $\sunny$: sunny, $\cloud$: cloudy, $\snow$: snowy).}
\setlength{\tabcolsep}{4pt} 
\centering\resizebox{0.49\textwidth}{!}{
\renewcommand{\arraystretch}{1.2}
\begin{tabular}{l|l|cccc}
\toprule \midrule
Sequence                              & Method      & \textbf{RTE [m] $\downarrow$}   & \textbf{RRE [$^\circ$] $\downarrow$} & \textbf{Success rate [\%] $\uparrow$} & \textbf{Runtime [s] $\downarrow$} \\ \midrule

\multirow{2}{*}{\makecell{\texttt{Mountain 01}\\\sunny}}  & Feat.~\cite{cen2018precise} + GICP  & 0.779                          & 1.5                                  & 99.6                                  & 0.342 \\
                                                          & Proposed          & \firstc \textbf{0.643}         & \firstc \textbf{0.9}                 & \firstc \textbf{100.0}                & \firstc \textbf{0.036}  \\ \midrule
                                      
\multirow{2}{*}{\makecell{\texttt{Mountain 02}\\\cloud}}  & Feat.~\cite{cen2018precise} + GICP & 1.018                          & 1.6                                  & 98.2                                  & 0.317 \\
                                                          & Proposed          & \firstc \textbf{0.881}         & \firstc \textbf{0.9}                 & \firstc \textbf{99.9}                 & \firstc \textbf{0.034} \\  \midrule
                                      
\multirow{2}{*}{\makecell{\texttt{Mountain 03}\\\snow}}   & Feat.~\cite{cen2018precise} + GICP & 0.915                          & 2.0                                  & 99.2                                  & 0.279 \\
                                                          & Proposed          & \firstc \textbf{0.778}         & \firstc \textbf{1.1}                 & \firstc \textbf{99.8}                 & \firstc \textbf{0.061} \\  \midrule
                                      
\multirow{2}{*}{\makecell{\texttt{Parking 01}\\\sunny}}   & Feat.~\cite{cen2018precise} + GICP & 0.409                          & 3.5                                  & 97.4                                  & 0.297 \\
                                                          & Proposed          & \firstc \textbf{0.370}         & \firstc \textbf{2.1}                 & \firstc \textbf{100.0}                & \firstc \textbf{0.063} \\  \midrule
\multirow{2}{*}{\makecell{\texttt{Parking 02}\\\sunny}}   & Feat.~\cite{cen2018precise} + GICP  & 0.603                          & 3.4                                  & 97.7                                  & 0.274 \\
                                                          & Proposed          & \firstc \textbf{0.570}         & \firstc \textbf{1.9}                 & \firstc \textbf{99.2}                 & \firstc \textbf{0.067} \\  \midrule
\multirow{2}{*}{\makecell{\texttt{Parking 03}\\\cloud}}   & Feat.~\cite{cen2018precise} + GICP  & \firstc \textbf{0.824}         & 4.6                                  & 93.4                                  & 0.283 \\
                                                          & Proposed          & 0.842                          & \firstc \textbf{2.9}                 & \firstc \textbf{100.0}                & \firstc \textbf{0.068} \\  \midrule
\multirow{2}{*}{\makecell{\texttt{Parking 04}\\\snow}}    & Feat.~\cite{cen2018precise} + GICP  & 0.927                          & 6.1                                  & 88.1                                  & 0.283 \\
                                                          & Proposed          & \firstc \textbf{0.698}         & \firstc \textbf{3.6}                 & \firstc \textbf{100.0}                & \firstc \textbf{0.070} \\  \midrule

\bottomrule
\end{tabular}}
\label{tab:regi_eval}
\vspace{-0.5cm}
\end{table}

\begin{figure}[t]
    \centering
    \def\width{0.49\textwidth}%
    \includegraphics[clip, trim={0 0 0 0}, width=0.49\textwidth]{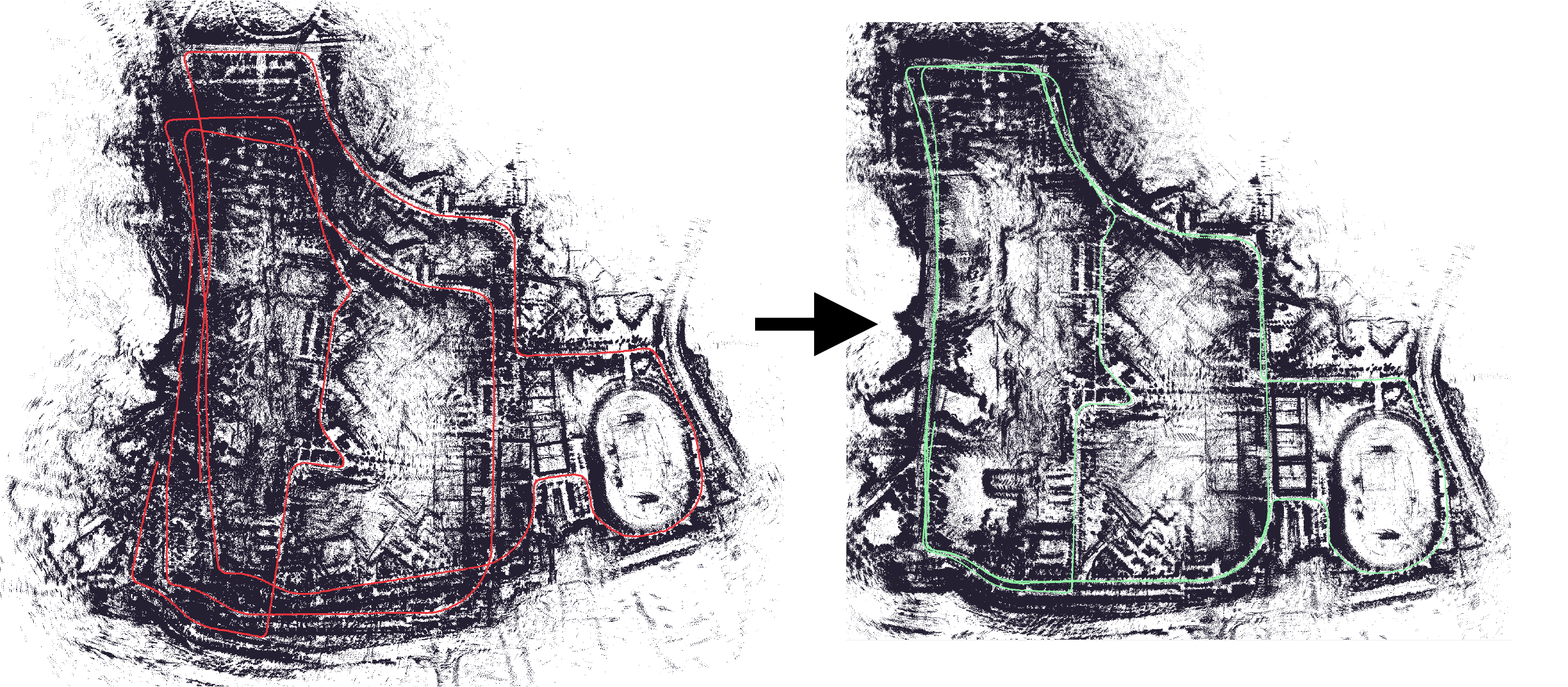}
    \caption{SLAM result on the \texttt{KAIST 03} sequence. (Left)~Odometry-only trajectory~(red) with accumulated drift. (Right)~Corrected trajectory~(green) after loop closure using our RadLoc, producing a globally consistent map.}
    \label{fig:slam} 
    \vspace{-0.5cm}
\end{figure}

\textbf{SLAM integration.}
One notable aspect of our RadLoc is its integration into a radar sensor-based SLAM pipeline to validate its practical applicability.
As illustrated in \figref{fig:slam}, the proposed method enables accurate global re-localization and consistent trajectory alignment across sessions.
These results confirm that our RadLoc not only performs well in retrieval metrics but also provides reliable global poses suitable for mapping and loop closure correction.

\begin{figure}[t]
    \centering
    \def\width{0.49\textwidth}%
    \includegraphics[clip, trim={0 0 0 0}, width=0.49\textwidth]{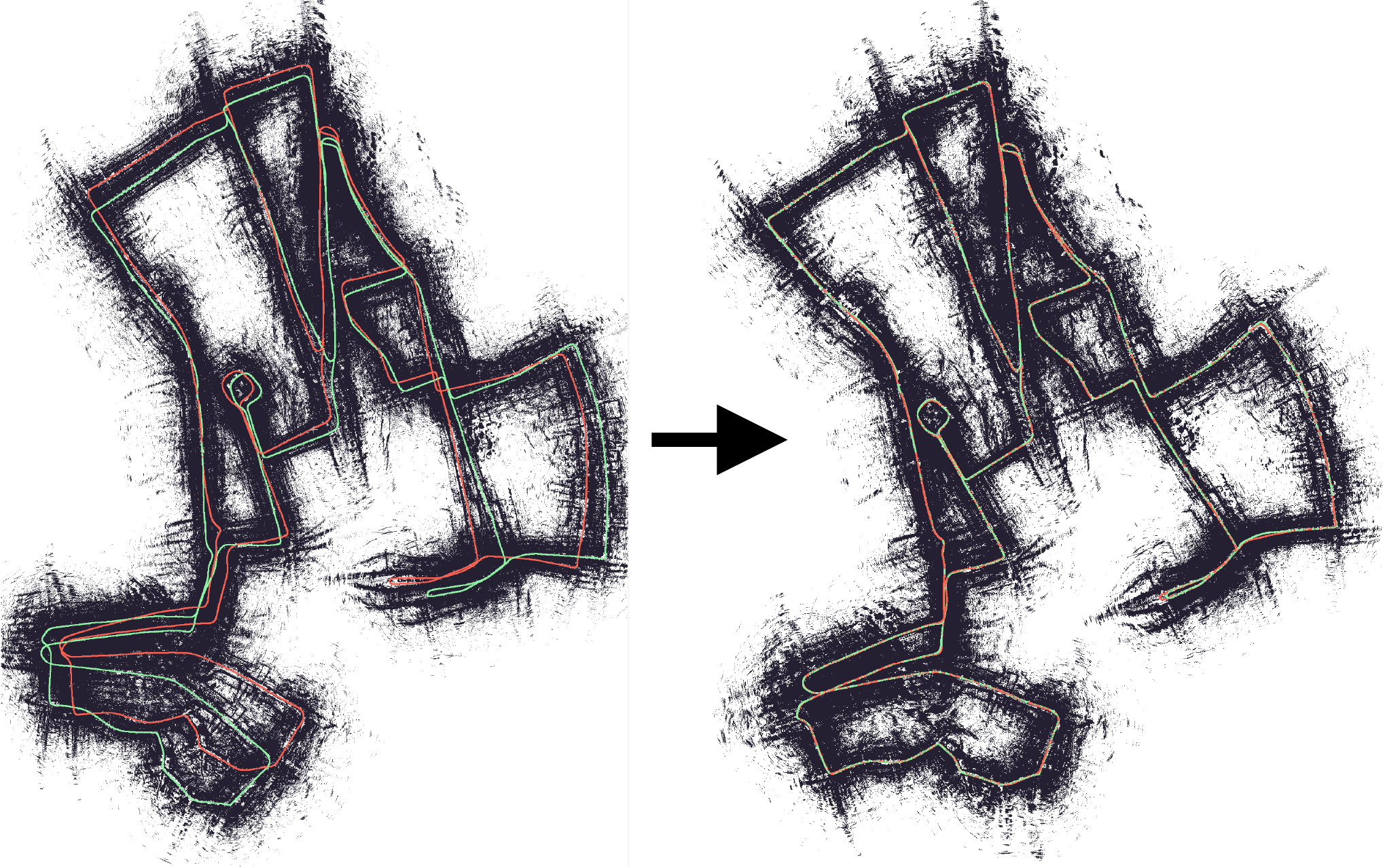}
    \caption{Multi-session SLAM result on the Oxford Radar Robotcar dataset~\cite{barnes2020oxford} using two sequences captured 8 days apart~(\texttt{19-01-10} and \texttt{19-01-18}). (Left)~Before multi-session alignment, showing misaligned trajectories. (Right)~After alignment using our RadLoc, we produced a globally consistent multi-session map.}
    \label{fig:mss} 
    \vspace{-0.5cm}
\end{figure}

\begin{figure}[t]
    \centering
    \def\width{0.49\textwidth}%
    \includegraphics[clip, trim={0 0 0 0}, width=0.49\textwidth]{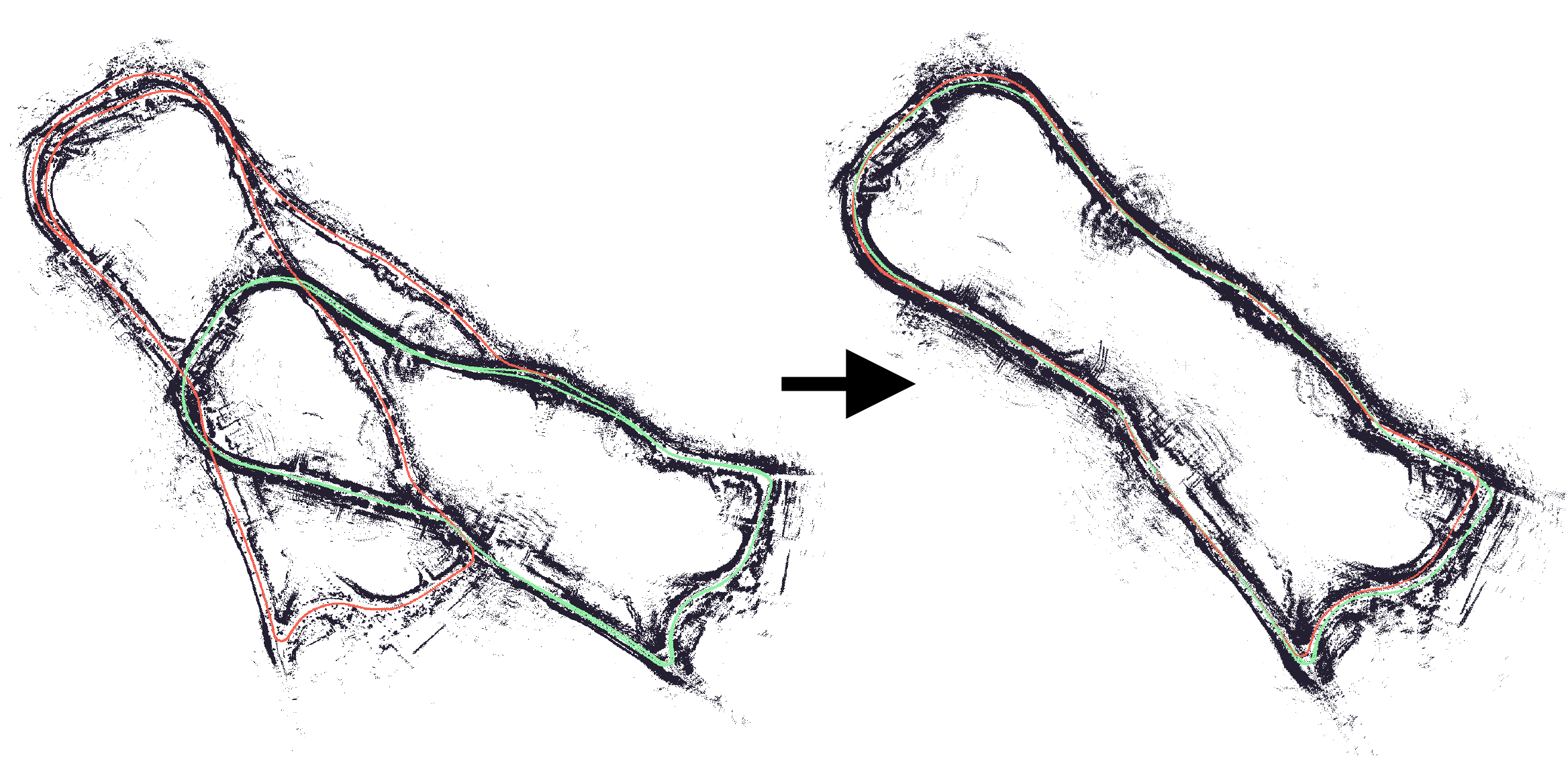}
    \caption{Cross-weather multi-session SLAM result on the Hercules dataset~\cite{kim2025hercules}. 
             Green and red trajectories represent sessions captured under sunny~(\sunny) and snowy~(\snow) conditions, respectively. 
             (Left)~Before multi-session alignment. (Right)~After alignment using our RadLoc, we demonstrate robust localization across substantially different weather conditions.}
    \label{fig:mmm} 
    \vspace{-0.2cm}
\end{figure}

\textbf{Multi-session and cross-weather alignment.}
In addition, we evaluate cross-session SLAM under environmental and weather changes.
As shown in \figref{fig:mss} and \figref{fig:mmm}, RadLoc successfully aligns trajectories acquired at different times and under different weather conditions.
Despite long-term structural changes and appearance variations, the proposed descriptor maintains sufficient distinctiveness for accurate global alignment, confirming robustness beyond single-session evaluation.

\textbf{Descriptor scalability.}
In multi-session map management, descriptor storage cost scales linearly with the number of sessions and map size, making compact representations essential for long-term deployment.
As shown in \tabref{tab:descriptor_size}, RadLoc maintains the smallest descriptor size among all compared methods, enabling efficient storage and retrieval even as the map grows across multiple sessions and weather conditions.
Note that this compactness is particularly beneficial for long-term mapping scenarios where memory and retrieval efficiency become critical.




\begin{figure}[t]
    \centering
    \def\width{0.49\textwidth}%
    \includegraphics[clip, trim={0 0 0 0}, width=0.49\textwidth]{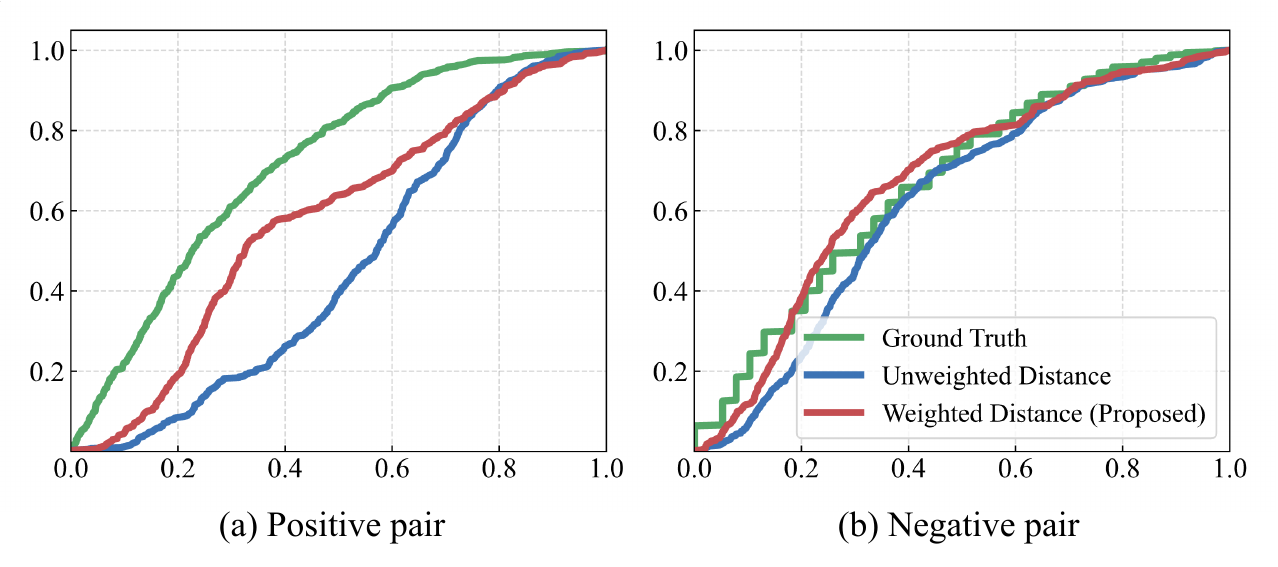}
    \caption{Kolmogorov-Smirnov plots comparing descriptor distance distributions against ground truth for (a)~positive and (b)~negative pairs. 
    \textit{Our weighted distance} achieves lower KS statistics than the unweighted variant for both positive~(\textit{0.2600} and 0.4880) and negative~(\textit{0.1260} and 0.1720) pairs. 
    All comparisons yield $p$-values below 0.005, confirming statistical significance. 
    All distributions, including ground truth, are normalized to a common scale for fair comparison.}
    \label{fig:ks_plot} 
    \vspace{-0.4cm}
\end{figure}
\begin{figure}[t]
    \centering
    \def\width{0.49\textwidth}%
    \includegraphics[clip, trim={0 0 0 0}, width=0.49\textwidth]{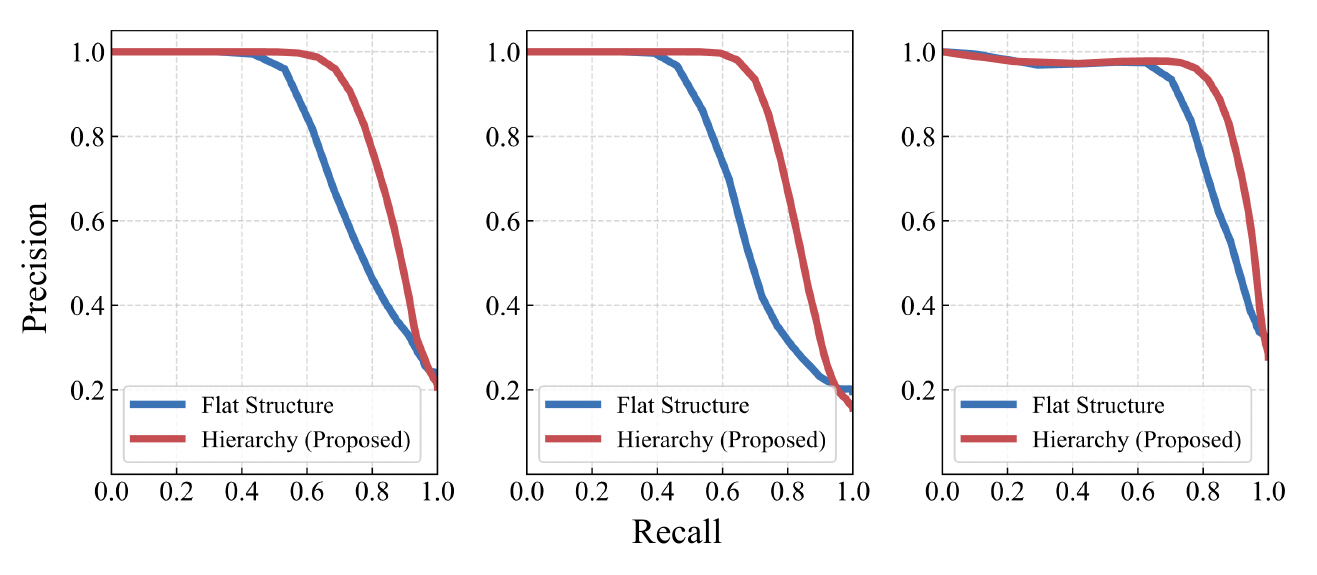}
    \vspace{-0.6cm}
    \caption{Precision-recall (PR) curves on three Oxford Radar Robotcar~\cite{barnes2020oxford} sequences~(\textit{i.e.,} \texttt{19-01-10}, \texttt{19-01-14}, \texttt{19-01-18}, from left to right) comparing KD-tree flat search using the full descriptor with our proposed hierarchical search~(Sec.~\ref{sec:place_recognition}).}
    \label{fig:abl_prcurve} 
    \vspace{-0.4cm}
\end{figure}
\begin{figure}[h!]
    \centering
    \def\width{0.49\textwidth}%
    \includegraphics[clip, trim={0 0 0 0}, width=0.49\textwidth]{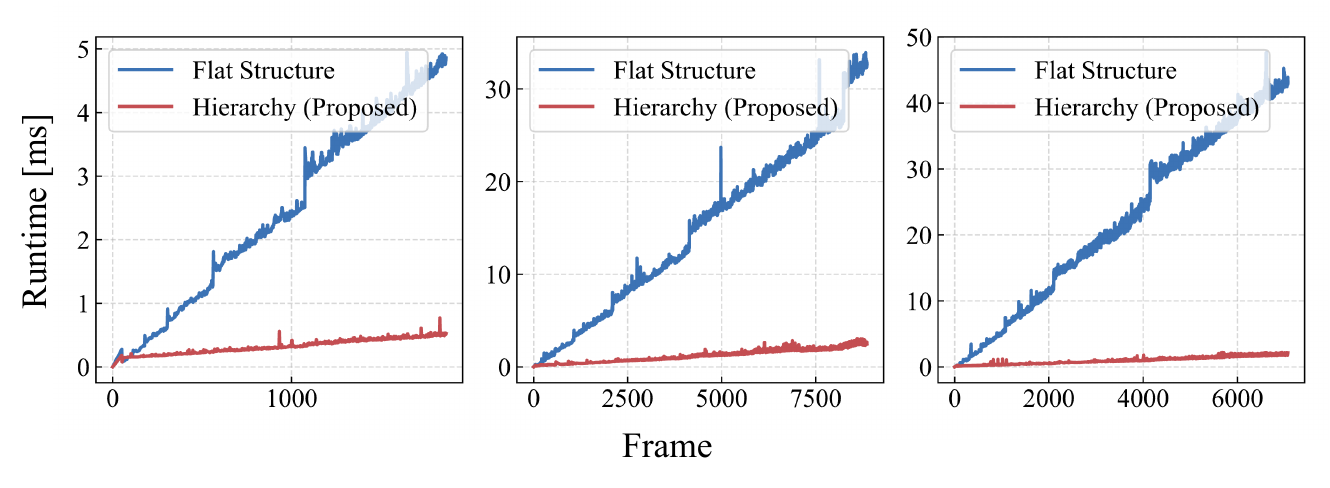}
    \vspace{-0.6cm}
    \caption{Retrieval runtime comparison between KD-tree flat search and our proposed hierarchical search on three datasets~(\textit{i.e.,} Hercules~\cite{kim2025hercules}, Oxford Radar Robotcar~\cite{barnes2020oxford}, OORD~\cite{gadd2024oord}, from left to right).
    Datasets are ordered by database size for a clearer illustration of scalability differences.}
    \label{fig:abl_runtime} 
    \vspace{-0.4cm}
\end{figure}

\subsection{Ablation Study}
\label{sec:ablation}
Finally, we conduct ablation studies to analyze the effects of the proposed range-dependent weighting and hierarchical retrieval strategies.

\textbf{Effect of range-dependent weighting.}
We examine how the proposed weighting affects the relationship between descriptor distances and ground truth spatial distances.
For both positive and negative pairs, we compare the distributions of ground truth distances and descriptor distances under weighted and unweighted formulations using the Kolmogorov–Smirnov~(KS) statistic.

As shown in \figref{fig:ks_plot}, the weighted formulation yields a closer alignment between descriptor distances and ground truth spatial separation for both positive and negative pairs.
In contrast, the unweighted descriptor exhibits stronger near-range bias, resulting in a less consistent correspondence with geometric distance.
We conclude that these results indicate the proposed weighting mitigates range-dependent bias while preserving consistency with underlying spatial relationships.

\begin{table}[t!]
\captionsetup{width=0.49\textwidth, justification=justified}
\caption{Descriptor size comparison~(unit: MB). RadLoc achieves the most compact representation, which is critical for scalable multi-session map management.}
\setlength{\tabcolsep}{4pt}
\centering\resizebox{0.49\textwidth}{!}{\normalsize
\renewcommand{\arraystretch}{1.2}
\begin{tabular}{l|ccccccccc}
\toprule \midrule
Method                 & Kidnapped & SHeRLoc & RSC          & RaPlace & RadVLAD & ReFeree         & Ours \\ \midrule
1st session (\sunny)   & 299.61    & 2.93    & \thirdc 2.74 & 6.59    & 599.22  & \secondc 0.78   & \firstc \textbf{0.73} \\ 
+ 2nd session (\snow)  & 199.70    & 1.95    & \thirdc 1.83 & 4.39    & 399.40  & \secondc 0.52   & \firstc \textbf{0.49} \\ \midrule 
Total [MB]             & 499.31    & 4.88    & \thirdc 4.57 & 10.98   & 998.62  & \secondc 1.30   & \firstc \textbf{1.22} \\
\midrule \bottomrule
\end{tabular}}
\label{tab:descriptor_size}
\vspace{-0.5cm}
\end{table}

\textbf{Effect of hierarchical retrieval.}
In addition, we evaluate the impact of the proposed hierarchical retrieval scheme.
Notably, as shown in \figref{fig:abl_prcurve}, hierarchical retrieval preserves recall performance comparable to exhaustive search.
At the same time, the runtime comparison in \figref{fig:abl_runtime} demonstrates substantial computational savings.
Note that this confirms the hierarchical strategy effectively reduces search complexity without degrading recognition quality.

\textbf{Effect of radial partition \textit{\textbf{K}}.}
Finally, we analyze the sensitivity of the partition index $K$, which determines the near-range segment boundary in the coarse retrieval stage.
As shown in \tabref{tab:abl_K}, a clear trade-off emerges: increasing $K$ consistently improves R@1 by leveraging more structural information, while AUC and F1 gradually decrease as far-range bins are incorporated into the weighted coarse stage.
Notably, this trend is consistent across both datasets despite differences in radar type and environment.
In practice, $K$ can be configured according to the application objective: a smaller $K$ favors higher precision suitable for safety-critical loop closure verification, while a larger $K$ prioritizes recall for broader candidate retrieval in exploratory scenarios.

\begin{table}[t!]
\captionsetup{width=0.49\textwidth, justification=justified}
\caption{Sensitivity analysis of the number of radial partitions~$K$ via place recognition performance. 
We evaluate four representative quartile-level configurations, \textit{i.e.,} $Q_1, Q_2, Q_3$, and $Q_4$. 
Note that we select {\setlength{\fboxsep}{1pt}\colorbox{mygray}{\textbf{$K$=20}}} in all experiments unless otherwise specified.}
\setlength{\tabcolsep}{4pt}
\centering\resizebox{0.49\textwidth}{!}
{\tiny
\renewcommand{\arraystretch}{1.2}
\begin{tabular}{l|cccccc}
\toprule \midrule
Sequence               & \multicolumn{3}{c}{\texttt{Mountain 03}}        & \multicolumn{3}{c}{\texttt{21-01-19}}\\ \cmidrule(lr){2-4} \cmidrule(lr){5-7} 
Evaluation             & R@1       & AUC     & F1                          & R@1       & AUC     & F1         \\ \midrule
$K$=8 ($Q_1$)            & \hspace{0.1cm} 0.934 \hspace{0.1cm}   & \hspace{0.1cm} 0.897 \hspace{0.1cm}    & \hspace{0.1cm} 0.803 \hspace{0.1cm}
                       & \hspace{0.1cm} 0.354 \hspace{0.1cm}   & \hspace{0.1cm} 0.654 \hspace{0.1cm}    & \hspace{0.1cm} 0.603 \hspace{0.1cm} \\ 
$K$=10 ($Q_2$)           & 0.946     & 0.901   & 0.800                       & 0.387     & 0.650   & 0.582  \\ 
\grayc \textbf{$K$=20 ($Q_3$)}           & \grayc \textbf{0.960}     & \grayc \textbf{0.893}   & \grayc \textbf{0.790}
                                       & \grayc \textbf{0.499}     & \grayc \textbf{0.582}   & \grayc \textbf{0.503} \\ 
$K$=40 ($Q_4$)           & 0.972     & 0.875   & 0.767                       & 0.551     & 0.546   & 0.459        \\ 
\midrule \bottomrule
\end{tabular}}
\label{tab:abl_K}
\vspace{-0.6cm}
\end{table}




\section{Conclusion}
In this paper, we presented \textit{RadLoc}, a fast, robust, and lightweight global localization pipeline for spinning radar. 
By replacing costly feature extraction with 1D CA-CFAR filtering and leveraging range-dependent descriptor design with hierarchical retrieval, our RadLoc forms a holistic end-to-end module that unifies descriptor generation, retrieval, and 3-DoF pose estimation. 
Extensive experiments on 15 sequences across 5 datasets demonstrated robust performance across diverse environments, weather conditions, and radar types, while achieving the smallest descriptor size and fastest retrieval time among state-of-the-art methods. 

In future work, we plan to extend our RadLoc to multi-robot SLAM, where its compact descriptor and lightweight pipeline are particularly advantageous under limited inter-robot communication bandwidth.

\small
\bibliographystyle{packages/IEEEtranN} 
\bibliography{packages/string-short, packages/references}

\end{document}